\def\year{2022}\relax
\documentclass[letterpaper]{article} 
\usepackage{aaai22}  
\usepackage{times}  
\usepackage{helvet}  
\usepackage{courier}  
\usepackage[hyphens]{url}  
\usepackage{graphicx} 
\graphicspath{{./figs/}}
\usepackage{natbib}  
\usepackage{caption} 
\DeclareCaptionStyle{ruled}{labelfont=normalfont,labelsep=colon,strut=off} 
\usepackage{algorithm}
\usepackage{algorithmic}
\usepackage{amssymb}
\usepackage{amsmath}
\newtheorem{defin}{Definition}
\usepackage{multirow}
\usepackage{booktabs}
\usepackage{newfloat}
\usepackage{listings}
\floatstyle{ruled}
\newfloat{listing}{tb}{lst}{}
\floatname{listing}{Listing}
\title{Is There More Pattern in Knowledge Graph? \\Exploring Proximity Pattern for Knowledge Graph Embedding}
\author{
    Ren Li\textsuperscript{\rm 1, 2}, 
    Yanan Cao\textsuperscript{\rm 1, 2}, 
    Qiannan Zhu\textsuperscript{\rm 3},
    Xiaoxue Li\textsuperscript{\rm 4},
    Fang Fang\textsuperscript{\rm 1, 2}
}
\affiliations{
    \textsuperscript{\rm 1}Institute of Information Engineering, Chinese Academy of Sciences \\
    \textsuperscript{\rm 2}School of Cyber Security, University of Chinese Academy of Sciences \\
    \textsuperscript{\rm 3}Gaoling School of Artificial Intelligence, Renmin University of China \\
    \textsuperscript{\rm 4}National Computer Network Emergency Response Technical Team/Coordination Center of China \\
    \{liren, caoyanan, fangfang0703\}@iie.ac.cn, 
    zhuqiannan@ruc.edu.cn, lixiaoxue93@outlook.com
}

\usepackage{bibentry}

\begin{document}

\maketitle

\begin{abstract}
Modeling of \textbf{relation pattern} is the core focus of previous Knowledge Graph Embedding works, which represents how one entity is \textit{related} to another semantically by some explicit relation.
However, there is a more natural and intuitive relevancy among entities being always ignored, which is that how one entity is \textit{close} to another semantically, without the consideration of any explicit relation. We name such semantic phenomenon in knowledge graph as \textbf{proximity pattern}.
In this work, we explore the problem of how to define and represent proximity pattern, and how it can be utilized to help knowledge graph embedding. Firstly, we define the proximity of any two entities according to their statistically shared queries, then we construct a derived graph structure and represent the proximity pattern from global view. Moreover, with the original knowledge graph, we design a \textbf{C}hained cou\textbf{P}le-\textbf{GNN} (\textbf{CP-GNN}) architecture to deeply merge the two patterns (graphs) together, which can encode a more comprehensive knowledge embedding.
Being evaluated on FB15k-237 and WN18RR datasets, CP-GNN achieves state-of-the-art results for Knowledge Graph Completion task, and can especially boost the modeling capacity for complex queries that contain multiple answer entities, proving the effectiveness of introduced proximity pattern.
\end{abstract}

\section{Introduction}
Knowledge Graphs (KGs) are large, graph-structured databases which store facts in triple form $(h, r, t)$, denoting that head entity $h$ and tail entity $t$ satisfy relation $r$.
Knowledge Graph Embedding (KGE) is a task of learning low-dimensional representation for entities and relations, so that the symbolic knowledge can be integrated into numerical model to support various down-stream knowledge based tasks, such as recommendation system \cite{CIKM_2018_Wang_RippleNet},  question answering \cite{Yasunaga_NAACL_2021_QA-GNN} and text generation \cite{ACL_2020_Zhang_Generation}, etc. To evaluate the effectiveness of learned embeddings, Knowledge Graph Completion (KGC) task is usually adapted \citep{AAAI_2018_Dettmers_ConvE_WN18RR,EMNLP_2019_Balazevic_TuckER,AAAI_2020_Vashishth_InteractE}, aiming at predicting tail entities $t$ given $(h, r, ?)$ or head entities $h$ given $(?, r, t)$. Essentially, KGC can be regarded as query-answer format and without loss of generality, we denote the query as $(h, r, ?)$ and the answer as $t$.

The general intuition of previous KGE works is to model the \textbf{relation pattern} of knowledge graph, which represents how one entity is \textit{related} to another semantically by some explicit relation. 
For example, TransE \citep{NeurIPS_2013_Bordes_TransE} models relations as addition operation from head entities to tail entities. RotatE \citep{ICLR_2019_Sun_RotatE} treats relations as rotation between entities on complex field. GNN-based models like R-GCN \citep{ESWC_2018_Schlichtkrull_R-GCN}, CompGCN \citep{ICLR_2020_Vashishth_CompGCN}, focus on capturing the relation pattern from global graph view, through neighborhood aggregation mechanism. 
While among entities, a more natural and intuitive relevancy is always ignored, which is that how one entity is \textit{close} to another semantically, without the consideration of any explicit relation. We name such semantic phenomenon in knowledge graph as \textbf{proximity pattern}.

Concretely, if a query $(h, r, ?)$ holds multiple answers $t_1\sim t_N$ simultaneously, we consider $t_1\sim t_N$ satisfy proximity pattern. We assume that answers under the same query share some common characteristics, which will close them semantically. 
For example, given query \texttt{(Robert Downey Jr.,} \texttt{act, ?)}, some of the answers like \texttt{The} \texttt{Avengers}, \texttt{Iron} \texttt{Man} \texttt{2}, \texttt{Avengers:} \texttt{Age of Ultron}, all possess the characteristics like superhero films, comic book films, product of Marvel Studios, to close them semantically. Other examples including movies from the same director, paper published from the same laboratory, multiple hyponyms of the same hypernym, support the same phenomenon.

\begin{figure}[t]
  \centering
  \includegraphics[width=\columnwidth]{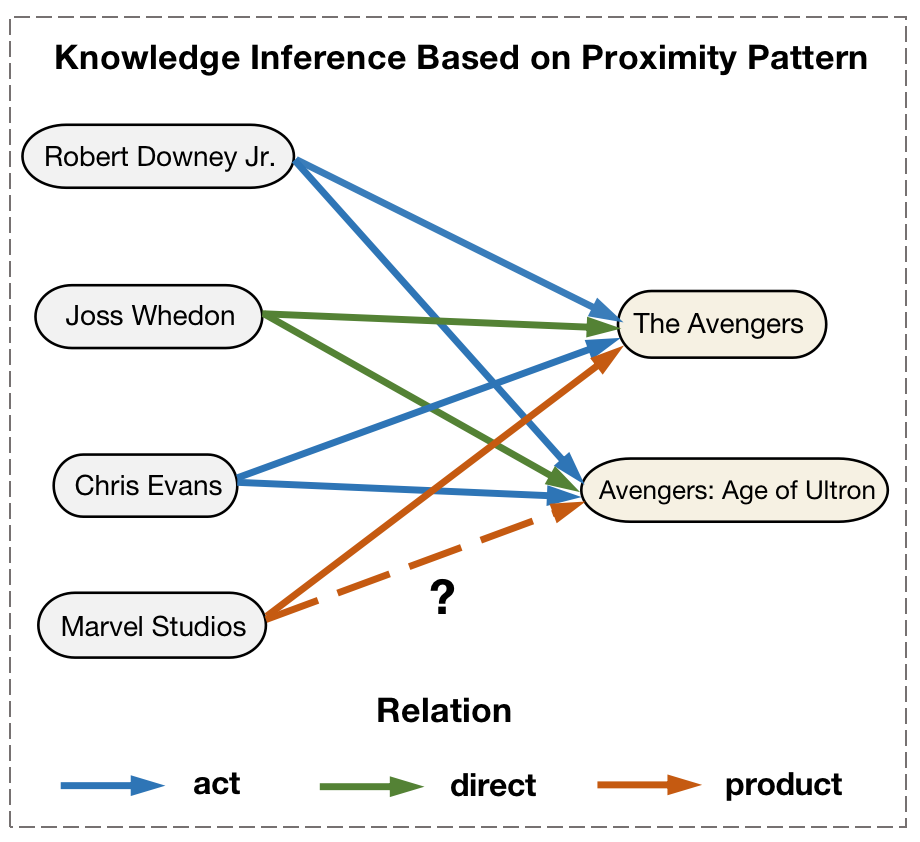}
  \caption{Knowledge inference based on proximity pattern. The high proximity between \texttt{The Avengers} and \texttt{Avengers:} \texttt{Age} \texttt{of} \texttt{Ultron} can be derived by large number of co-occurrences under the same query. Then if we have \texttt{(Marvel Studios,} \texttt{product,} \texttt{The Avengers)} during training, a strong confidence will be given to the hold of the fact \texttt{(Marvel Studios,} \texttt{product,} \texttt{Avengers:} \texttt{Age of Ultron)}.}
  \label{fig: proximity_inference}
\end{figure}

Intuitively, it is necessary to capture and model the proximity pattern since it can provide beneficial evidences during inference phase. Like in Figure \ref{fig: proximity_inference}, we can know that \texttt{The Avengers} has strong semantic proximity with \texttt{Avengers:} \texttt{Age} \texttt{of} \texttt{Ultron}, because of their frequent ``co-answer''s under the same query. Then if there is fact \texttt{(Marvel Studios,} \texttt{product,} \texttt{The Avengers)} during training, it will be easy to infer out \texttt{(Marvel Studios,} \texttt{product,} \texttt{Avengers:} \texttt{Age of Ultron)}.
Note that though for some GNN-based KGE works, neighbor aggregation mechanism can also capture the proximity information to some extent, because of no explicit guidance and attendance, the effective utilization is still limited. 
 
\begin{figure*}
  \centering
  \includegraphics[width=\textwidth]{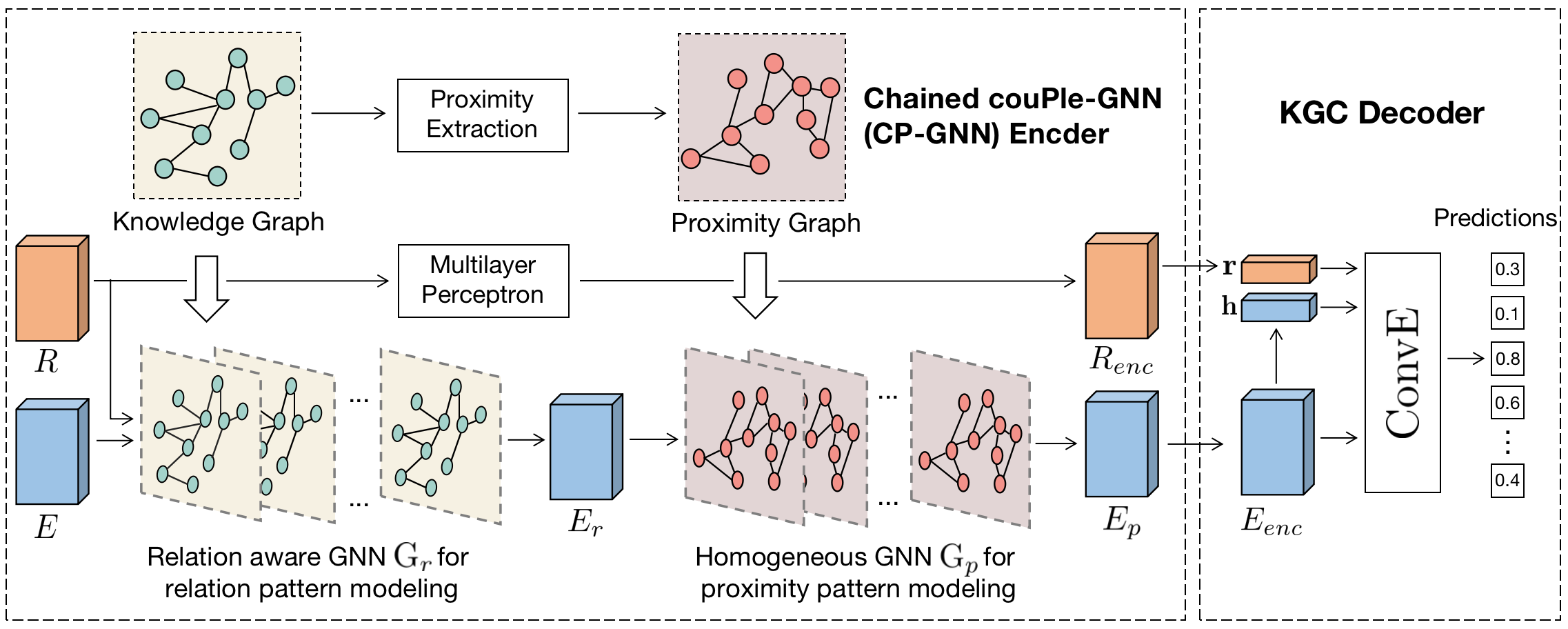}
  \caption{The overall architecture of CP-GNN and its application to downstream Knowledge Graph Completion (KGC) task. The process follows the encoder-decoder paradigm. During encoding step, the initial entity embedding $E$ is sequently modeled by a relation aware GNN $\mathrm{G}_r$ on knowledge graph and a homogeneous GNN $\mathrm{G}_p$ on proximity graph. The initial relation embedding $R$ is transformed by an Multilayer Perceptron (MLP). During decoding step, ConvE \citep{AAAI_2018_Dettmers_ConvE_WN18RR} is chose as the decoder to perform the KGC task.
  }
  \label{fig: model}
\end{figure*}

Inspired by this intuition, in this work we explore the problem of how to define and represent proximity pattern, and how it can be utilized to help knowledge graph embedding. The overall framework is demonstrated in Figure \ref{fig: model}.
Firstly, we define the proximity of any two entities according to the statistically shared query between them, then based on the proximity value we construct a derived proximity graph. This can be seen that we extract a new semantic graph structure from the original knowledge graph, which can represent the proximity pattern from a global view. 
Then we employ Graph Neural Networks (GNNs) to model and merge two patterns (graphs) together, as their powerful graph structure learning ability revealed in recent years \cite{ICLR_2017_Kipf_GCN,ICLR_2018_Velickovic_GAT,ICLR_2019_Xu_GIN}.
We first employ a relation aware GNN on knowledge graph and a homogeneous GNN on proximity graph respectively, to capture the global semantic interactions within pattern. Then we stack the two GNNs together, to capture the deep semantic interactions across pattern. 
We name the overall architecture as \textbf{C}hained cou\textbf{P}le-\textbf{GNN} (\textbf{CP-GNN}), to represent the separate and sequence modeling characteristics applied here. In CP-GNN, GNN $\mathrm{G}_p$ is put back to the $\mathrm{G}_r$, because we tend to serve proximity modeling as an enhancement of original knowledge graph embedding. After fusing into the proximity pattern, the encoder will obtain a more comprehensive knowledge embedding. Finally, ConvE \citep{AAAI_2018_Dettmers_ConvE_WN18RR} is chose as the decoder to perform the Knowledge Graph Completion task.

In summary, our main contributions are as follows:
\begin{itemize}
  \item To our best knowledge, this is the first work to propose proximity pattern concept in knowledge graph field, which serves as a different semantic assumption compared with traditional relation pattern. 
  \item We dive into the way to fuse proximity pattern and relation pattern for more comprehensive knowledge graph embedding, and design a \textbf{C}hained cou\textbf{P}le-GNN (\textbf{CP-GNN}) architecture to sufficiently capture the global and deep interactions of two patterns.
  \item Extensive experiments on FB15k-237 and WN18RR datasets demonstrate the effectiveness of our proposed proximity pattern and CP-GNN KGE model.
\end{itemize}

\section{Related Work}
Knowledge Graph Embedding (KGE) is an active research area, where literature works mainly aim to model the relation pattern of KG and can be roughly divided into three families \cite{TKDE_2017_Wang_Survey, 2020_Arora_KG-GNN}: 
\textbf{Translational Distance Models} apply distance-based scoring function and model relations as some operation between head and tail entity, like addition operation in TransE \citep{NeurIPS_2013_Bordes_TransE}, hyper-plane addition in TransH \citep{AAAI_2014_Wang_TransH}, complex field rotation in RotatE \citep{ICLR_2019_Sun_RotatE}, etc. 
\textbf{Semantic Matching Models} employ similarity-based function to directly model the relation into the triple. DistMult \citep{ICLR_2015_Yang_DistMult}, ComplEx \citep{ICML_2016_Trouillon_ComplEx} employ multiplication model to represent the likelihood of a fact. ConvE \citep{AAAI_2018_Dettmers_ConvE_WN18RR}, InteracE \citep{AAAI_2020_Vashishth_InteractE} applies neural networks for similarity modeling. 
\textbf{GNN-based Models} are proposed to encode relation pattern from a global graph structure angle. R-GCN \citep{ESWC_2018_Schlichtkrull_R-GCN} improves GCNs by introducing relation-specific transformation during neighbor aggregation. A2N \citep{ACL_2019_Bansal_A2N} introduces an attentional aggregating mechanism to adaptively merge relevant neighborhood into query representation. Moreover, Comp-GCN \citep{ICLR_2020_Vashishth_CompGCN} generalizes different neighbor aggregation methods of knowledge graph as entity-relation composition operations, giving a unified framework GNN-based works. All three families of KGE works lack the consideration of latent semantic relevancy that describes how entities are close to each other without explicit relation. 

\section{Methodology}
In section \ref{sec: problem}, we firstly introduce the Knowledge Graph Embedding problem. Then in section \ref{sec: pp}, we introduce the definition, representation and modeling method of proximity pattern, and in section \ref{sec: rp}, we discuss the modeling method of relation pattern. Finally in section \ref{sec: CP-GNN}, we will introduce the CP-GNN architecture and the model training process. 

\subsection{Problem Definition}
\label{sec: problem}
A knowledge graph is denoted by $\mathcal{G}=(\mathcal{E}, \mathcal{R}, \mathcal{F})$, where $\mathcal{E}$ and $\mathcal{R}$ represents the set of entities and relations, and $\mathcal{F} = \{(h_i, r_i, t_i)\} \subseteq \mathcal{E} \times \mathcal{R} \times \mathcal{E}$ is the set of triple facts. Knowledge Graph Embedding (KGE) task tries to learn a function $\phi: \mathcal{E} \times \mathcal{R} \times \mathcal{E} \mapsto \mathbb{R}$ such that given a query $q = (h, r, ?)$ and an entity $t$, the output score $\phi(q, t) \in \mathbb{R}$ measures the likelihood of $t$ being one of the answers of $q$.

During calculation, $\mathcal{E}$ and $\mathcal{R}$ will be represented as embedding matrix. Entity embedding matrix is formulated as $E \in \mathbb{R}^{n_e \times d}$, where $n_e$ denotes the number of entities, and $d \ll n_e$ is the dimension of embedding. For single entity $e_i$, we denote its corresponding vector as boldface form $\mathbf{e}_i \in \mathbb{R}^{d}$, which is the transposition of $i$-th row of $E$. Similarly, relation embedding matrix is denoted as $R \in \mathbb{R}^{n_r \times d}$, where $n_r$ is the number of relations and $\mathbf{r}_i$ the embedding vector of relation $r_i$. 

\subsection{Proximity Pattern Modeling}
\label{sec: pp}

\subsubsection{Proximity Pattern Definition}
In order to draw forth the proximity concept, we firstly give the definition of Query-Answer (QA) pair: 
\begin{defin}[Query-Answer pair, QA pair]
  A Query-Answer (QA) pair $(q, a)$ consists of a query $q=(h, r, ?)$ or $(?, r, t)$, and an answer set $a=\{e_1, ..., e_m\}$ that represents all the answer entities of query $q$ in knowledge graph.  
  \label{def: QA pair}
\end{defin}
Each triple fact $(h, r, t)$ is included in two QA pairs, for $(h, r, ?)$ and $(?, r, t)$ direction respectively. From the triple set $\mathcal{F} = \{(h_i, r_i, t_i)\}$, we can obtain a QA pair set $\mathcal{T} = \{(q_k, a_k)\}$.

The core hypothesis we put forward for the \textit{proximity pattern} is that entities in the same answer set share characteristics, like the movies from same director, the paper published from same laboratory, etc. The term \textit{proximity} means that such entities should be close to each other both in cognitive semantic space and numerical embedding space. To formally describe the concept, we propose \textit{Proximity Measure} (PM) as following:
\begin{defin}[Proximity Measure, PM]
  Given a QA pair $(q_k, a_k)$ with $|a_k| > 1$, for $\forall e_i, e_j \in a_k \, (i \neq j)$, the Proximity Measure (PM) between $e_i, e_j$ with regard to $(q_k, a_k)$ is defined as: $p^k_{ij}=p^k_{ji}=\frac{\mathrm{max}(M-|a_k|, 0)}{M-2}$.
  \label{def: PM}
\end{defin}
$M\ge 2$ is a hyper-parameter that represents the threshold size of the answer set. PM is inversely proportional to the answer set size, which takes the maximum value of $1$ for $|a_k|=2$, and minimum value of $0$ for $|a_k|>=M$. This comes from the observation that the concentration extent of semantic proximity decays when the number of answers increases. For example, the more movies one actor participates in, lower the probability that these movies share same characteristics (genre, style, etc.) will be. 

Beyond single particular query, by considering all the observed shared queries in dataset, we define Statistical Proximity Measure as: 
\begin{defin}[Statistical Proximity Measure, SPM]
  For any entity $e_i, e_j \in \mathcal{E}$, the Statistical Proximity Measure(SPM) between $e_i$ and $e_j$ is: 
  $$
  p_{ij} = p_{ji} = \sum_{k=1}^{|\mathcal{T}|} p^k_{ij}
  $$
  where $\mathcal{T}=\{(q_k, a_k)\}$ is the set of QA pair, and $p^k_{ij}=0$ if $e_i$ or $e_j$ is not in $a_k$.
  \label{def: SPM}
\end{defin}

After obtaining the SPM between any entity pair, we can give the formal definition of the proximity pattern of knowledge graph.

\begin{defin}[Proximity Pattern]
  For a knowledge graph $\mathcal{G}=(\mathcal{E}, \mathcal{R}, \mathcal{F})$, the Proximity Pattern of $\mathcal{G}$ is a matrix: 
  $$
  P^{\mathcal{G}} \in \mathcal{R}^{n_e \times n_e}, [P^{\mathcal{G}}]_{ij} = p_{ij}
  $$
  where $n_e$ is the number of entities in knowledge graph, and $[P^{\mathcal{G}}]_{ij}$ is the value of ij-th entry of $P^{{\mathcal{G}}}$.
  \label{def: PP}
\end{defin}

\subsubsection{Proximity Graph Construction}
\begin{figure}[t]
  \centering
  \includegraphics[width=\columnwidth]{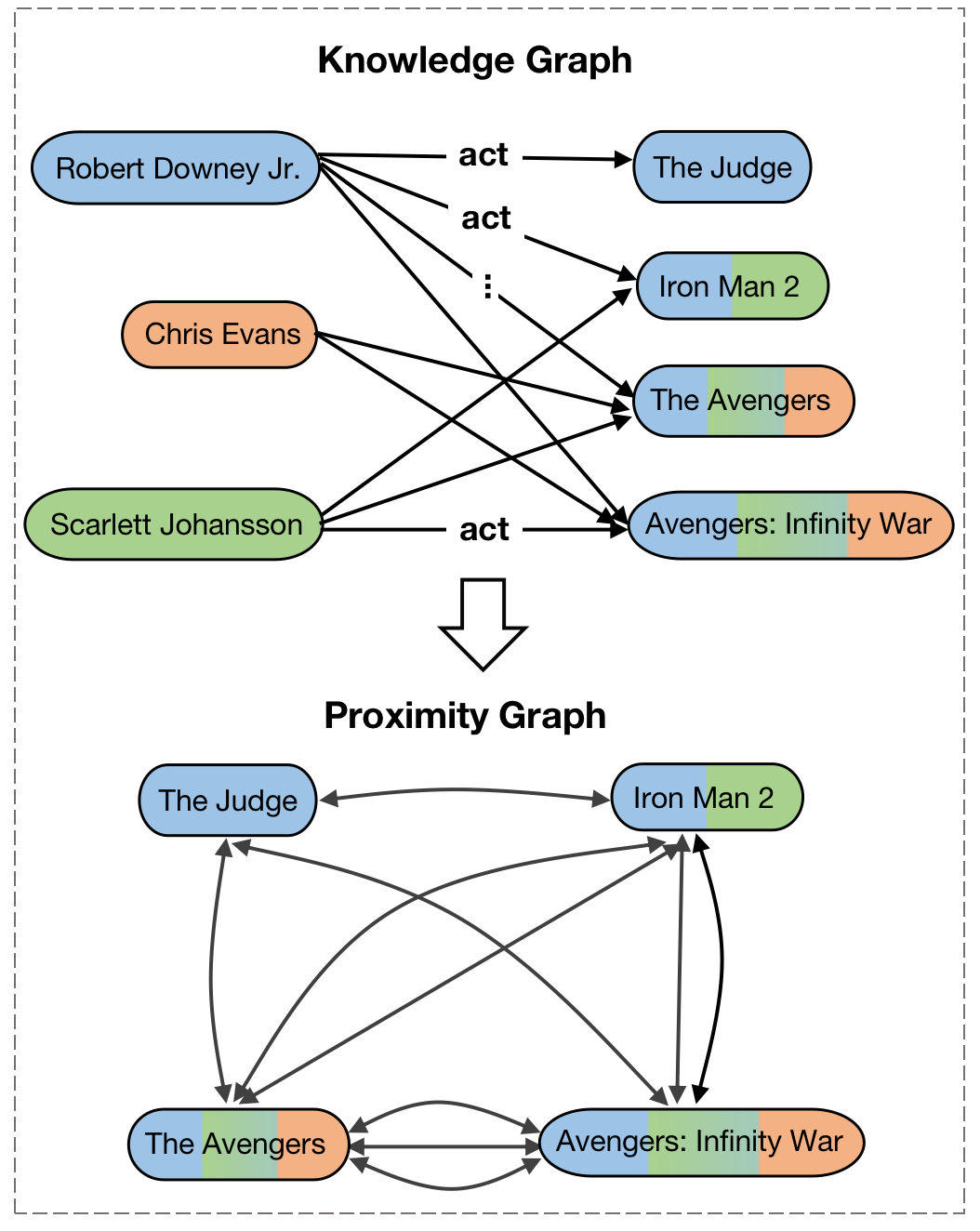}
  \caption{Proximity graph extraction process. For clarity we only select \texttt{act} relation cases. The colors of tail node denote the queries they are involved in, so if there is one kind of common color for two tail nodes, one edge should be connected between them in proximity graph. }
  \label{fig: proximity_extraction}
\end{figure}

According to the Definition \ref{def: PP}, proximity pattern is a matrix $P^{\mathcal{G}}$ that describes semantic proximity extent between any entity pair. In order to represent the global interactions among entities, we construct a proximity graph based on $P^{\mathcal{G}}$. Specifically, given a minimum threshold $I$, if the entry $[P^{\mathcal{G}}]_{ij} > I$, we will connect an undirected edge between $e_i$ and $e_j$ in graph, and set the weight of edge to $[P^{\mathcal{G}}]_{ij}$. The illustration can be seen in figure \ref{fig: proximity_extraction}. 

\subsubsection{GNN for Modeling Proximity Pattern}
After constructing the proximity graph, we can utilize extensive graph representation learning works to encode proximity pattern into embeddings. Here we choose the tool of Graph Neural Network (GNN), which has shown the competence for modeling global and complex graph features.

We start with a single GNN layer. For each entity $e_i$, denoting its input entity embedding for $l$-th layer as $\mathbf{e}_i^{(l)}$, the neighbor aggregation mechanism is formulated as:
\begin{align}
  \mathbf{n}_i^{(l)} &= \sum_{e_j\in \mathcal{N}_i} \alpha_{ij} \, \mathbf{e}_j^{(l)} \\
  \alpha_{ij} &= \frac{\mathrm{exp}([P^{\mathcal{G}}]_{ij})}{\sum_{e_z \in \mathcal{N}_i} \mathrm{exp}([P^{\mathcal{G}}]_{iz})}
  \label{eq: pp_agg}
\end{align}
$\mathcal{N}_i$ denotes the neighborhood of entity $e_i$ in proximity graph. $\alpha_{ij}$ is the normalization version of $[P^{\mathcal{G}}]_{ij}$ computed by $\mathrm{Softmax}$ function across $\mathcal{N}_i$. $\mathbf{n}_i^{(l)}$ is the neighbor representation of entity $e_i$ on $l$-th layer, which can be seen as the weighted summarization of neighbors according to their proximity extent with $e_i$.

After getting the neighbor representation, we employ it to update the entity embedding:
\begin{equation}
  \mathbf{e}_i^{(l+1)} = \sigma(W_p^{(l)} \mathbf{n}_i^{(l)}) + \mathbf{e}_i^{(l)}
  \label{eq: pp_upd}
\end{equation}
$W_p^{(l)} \in \mathbb{R}^{d \times d}$ is the linear transformation matrix for proximity pattern. $\mathbf{e}_i^{(l+1)}$ is either the output of $l$-th GNN layer or the input of $(l+1)$-th layer. $\sigma$ is the non-linear activation function.

The modeling process corresponds to $G_p$ in figure \ref{fig: model}. After $L$ layers' aggregation and updating, we serve the resulted entity embedding matrix $E^{(L)}$ as the output of $\mathrm{G}_p$ and denote as $E_{p}$, into which the proximity pattern is integrated. 

\subsection{Relation Pattern Modeling}
\label{sec: rp}
The relation pattern describes how one entity is related to another by some explicit relation, which is represented by knowledge graph triples, and can be formally defined as: 

\begin{defin}[Relation Pattern]
  For a knowledge graph $\mathcal{G}=(\mathcal{E}, \mathcal{R}, \mathcal{F})$, the Relation Pattern of $\mathcal{G}$ is a tensor: $R^{\mathcal{G}} \in \mathcal{R}^{n_e \times n_e \times n_r}$, 
  \begin{equation*}
    [R^{\mathcal{G}}]_{ijk} = 
    \begin{cases}
      1, & (e_i, r_k, e_j) \in \mathcal{F} \\
      0, & otherwise
    \end{cases}
  \end{equation*}
  where $n_e$ and $n_r$ is the number of entities and relations in knowledge graph, and $[R^{\mathcal{G}}]_{ijk}$ is the value of ijk-th entry of $R^{{\mathcal{G}}}$.
  \label{def: RP}
\end{defin}

According to Definition \ref{def: RP}, the knowledge graph itself is the sparse embodiment of the relation pattern, hence to capture relation pattern in a global way, we introduce a relation-aware GNN model, which corresponds to $\mathrm{G}_r$ in figure \ref{fig: model}. 

For a single layer, the neighbor aggregation function is formulated as: 
\begin{equation}
  \mathbf{n}_i^{(l)} = \sum_{(e_j, r_j)\in \mathcal{N}_i} \alpha_{ij}^{(l)} \, \varphi(\mathbf{e}_j^{(l)}, \mathbf{r}_j)
  \label{eq: rp_agg}
\end{equation}
$\mathcal{N}_i = \{(e_j, r_j)|(e_j, r_j, e_i) \in \mathcal{F} \}$ denotes the \textit{relational neighbors} of entity $e_i$, which are the neighbor entities associated with the connecting relation.
$\mathbf{r}_j$ is the initial embedding of relation and remains invariable for every layer, which will be explained later.  $\varphi(\mathbf{e}, \mathbf{r})$ is the composition function to fuse the entity and relation information. The selection includes additive function: $\varphi(\mathbf{e}, \mathbf{r}) = \mathbf{e} + \mathbf{r}$; multiplication function: $\varphi(\mathbf{e}, \mathbf{r}) = \mathbf{e} * \mathbf{r}$, where $*$ denotes element-wise multiplication; Multilayer Perceptron: $\varphi(\mathbf{e}, \mathbf{r}) = \mathrm{MLP}([\mathbf{e} || \mathbf{r}])$, where $||$ is the vector concatenation operation. Here we choose additive function, which is the most computational effective one. $\alpha_{ij}^{(l)}$ is the weight for relational neighbor $(e_j, r_j)$, the discussion of different weight choices is placed in Appendix \ref{app: attention}.

Then we update the entity embedding by:
\begin{equation}
  \mathbf{e}_i^{(l+1)} = \sigma(W_r^{(l)} \mathbf{n}_i^{(l)}) + \mathbf{e}_i^{(l)}
  \label{eq: rp_upd}
\end{equation}
$W_{r}^{(l)} \in \mathbb{R}^{d \times d}$ is the linear transformation matrix for relation pattern. After $L$ layers, we use the output entity embedding matrix $E^{(L)}$ as the final output of $\mathrm{G}_r$, denoted as $E_{r}$.

For equation \ref{eq: rp_agg}, different from previous works \citep{ICLR_2020_Vashishth_CompGCN} where relations are also updated in each layer, we remain the relation embedding the same. Here we assume that relations serve as the translation operation between entities, and the operation itself should keep consistent effect in different layers. 

\subsection{Pattern Fusion and Training}
\label{sec: CP-GNN}
In this section we will introduce our proposed CP-GNN framework to merge two patterns together and an end-to-end encoder-decoder training paradigm.

At encoding stage, as introduced in section \ref{sec: pp} and \ref{sec: rp}, two GNNs are employed to capture the global semantic interactions within pattern, which denoted as $\mathrm{G}_p$ and $\mathrm{G}_r$. 
Then we stack two GNNs together to capture the deep semantic interactions across pattern. GNN $\mathrm{G}_p$ is put back to the $\mathrm{G}_r$, because we tend to serve proximity modeling as an enhancement of original knowledge graph embedding. After fusing into the proximity pattern, the encoder will obtain a more comprehensive knowledge embedding.

Given the initial entity and relation embedding matrix as $E \in \mathbb{R}^{n_e \times d} $ and $R \in \mathbb{R}^{n_r \times d}$, the encoder process follows:
\begin{align}
  E_r &= \mathrm{G}_r(E, R) \\
  E_p &= \mathrm{G}_p(E_r)
\end{align}
For relation embedding $R$, we transform it by an Multilayer Perceptron (MLP) to get the output embedding. The total output of the encoder can be described as: 
\begin{align}
  E_{enc} &= E_p \\
  R_{enc} &= \mathrm{MLP}(R)
\end{align}

In decoder module, we perform the task of Knowledge Graph Completion. The module takes the embedding pair of query $\mathbf{q}=(\mathbf{h}, \mathbf{r})$ as input, and aims to measure the answer likelihood with regard to all the entities $\{\mathbf{e}_1, ..., \mathbf{e}_{n_e}\}$ of $E_{enc}$. Here we choose ConvE \citep{AAAI_2018_Dettmers_ConvE_WN18RR} as our decoder implementation, which uses 2D convolutional neural network to match query and answers. We refer readers to original paper for more details, and here we directly utilize the decoder function as:
\begin{equation}
  \mathrm{ConvE}(\mathbf{q}, E_{enc}) = \mathbf{o}
\end{equation} 
$\mathbf{o} \in \mathbb{R}^{n_e}$ is the matching scores between query and all entities. Then we use binary cross entropy loss to measure the difference between model output $\mathbf{o}$ and target $\mathbf{t} \in \mathbb{R}^{n_e}$: 
\begin{equation}
  \begin{aligned}
    \mathcal{L}(\mathbf{o}, \mathbf{t})=-\frac{1}{n_e} \sum_{i}^{n_e}\mathbf{t}[i] \cdot \log (\mathbf{o}[i]) + \\ 
    (1-\mathbf{t}[i]) \cdot \log (1-\mathbf{o}[i])
  \end{aligned}
\end{equation}
$\mathbf{t}[i]$ and $\mathbf{o}[i]$ denote the $i$-th entry of the vector. Then Stochastic Gradient Descent (SGD) algorithm is applied to train model predictions approximating targets.

\section{Experiments}

\subsection{Experiment Setup}
\label{sec: exp setup}
\paragraph{Dataset}
We conduct experiments of Knowledge Graph Completion task on two commonly used public datasets: FB15k-237 \citep{2015_Toutanova_FB15k-237} and WN18RR \citep{AAAI_2018_Dettmers_ConvE_WN18RR}. FB15k-237 contains entities and relations from Freebase, which is a large commonsense knowledge base. WN18RR is created from WordNet, a lexical database of semantic relations between words. Each dataset is split into train, valid and test sets. The statistics of two dataset are summarized at Table \ref{tab: dataset statistics}.

\paragraph{Evaluation Protocol}
The model performance is measured by five frequently used metrics: MRR (the Mean Reciprocal Rank of correct entities), MR (the Mean Rank of correct entities), Hits@1, Hits@3, Hits@10 (the accuracy of correct entities ranking in top 1/3/10). We follow the filtered setting protocol \citep{NeurIPS_2013_Bordes_TransE} for evaluation, i.e. all the other true entities appearing in train, valid and test set are excluded when ranking. In addition, based on the observation of \citep{ACL_2020_Sun_re-eval}, to eliminate the problem of abnormal score distribution, if prediction target have the same score with multiple other entities, we take the average of upper bound and lower bound as result.

\paragraph{Knowledge Graph Construction}
Here we give some details for knowledge graph construction process, which are shown effective during our experiments:
\begin{itemize}
  \item Following the \citep{ICLR_2020_Vashishth_CompGCN} work, we transform the knowledge graph to undirected graph, by introducing an inverse edge $(t, r^{-1}, h)$ for each edge $(h, r, t)$, which aims to pass the information bidirectionally and enhance the graph connectivity.
  \item For each training batch, we randomly remove a proportion of edges in the knowledge graph. So that the model will focus more on how to predict missing edges in the graph, which is closer to the inference process.
\end{itemize}

\paragraph{Hyper-parameter setting}
The hyper-parameters involved in our work include: batch size from $\{256, 512, 1024\}$, learning rate from $\{1e^{-4}, 3e^{-4}, 5e^{-3}\}$, dimension of entity and relation embedding from $\{500, 1000\}$, layer number of knowledge graph from $\{1, 2, 3\}$, layer number of proximity graph from $\{1, 2, 3\}$, randomly removing rate of batch edges from $\{0.1, 0.3, 0.5, 0.7, 1.0\}$, maximum set size $M$ of QA pair from $\{25, 50, 100, 500\}$ (Definition \ref{def: PM}), minimum SPM threshold $I$ from $\{0.5, 1, 3, 5\}$ (Definition \ref{def: SPM}). We tune the hyper-parameters by grid search algorithm. 

\begin{table}[t]
  \centering
  \begin{tabular}{lcc}
    \toprule
    \textbf{Dataset} & \textbf{FB15k-237} & \textbf{WN18RR}\\
    \hline\hline
    \# entity          & 14,541     & 40,943  \\
    \# relation        & 237        & 11      \\
    \# train triple    & 272,115    & 86,835  \\
    \# valid triple    & 17,535     & 3,034   \\
    \# test triple     & 20,466     & 3,134   \\
    \bottomrule
  \end{tabular}
  \caption{Dataset statistics}
  \label{tab: dataset statistics}
\end{table}

\begin{table*}
  \renewcommand\arraystretch{1.1}
  \centering
  \begin{tabular}{lccccccccccc}
    \toprule
    \multirow{2}{*}{\textbf{Models}} & \multicolumn{5}{c}{\textbf{FB15k-237}} & & \multicolumn{5}{c}{\textbf{WN18RR}} \\
    \cline{2-6} \cline{8-12}
    & MRR & MR & H@1 & H@3 & H@10 & & MRR & MR & H@1 & H@3 & H@10 \\
    \hline \hline
    \multicolumn{12}{l}{\textbf{Translational Distance}} \\
    TransE \cite{NeurIPS_2013_Bordes_TransE}     & .294 & 357 & - & - & .465 & & .226 & 3384 & - & - & .501 \\
    RotatE \cite{ICLR_2019_Sun_RotatE}      & .338 & 177 & .241 & .375 & .533 & & .476 & 3340 & .428 & .492 & \textbf{.571} \\
    PaiRE \cite{ACL_2021_Chao_PaiRE}        & .351 & \textbf{160} & .256 & .387 & .544 & & - & - & - & - & - \\
    \hline
    \multicolumn{12}{l}{\textbf{Semantic Matching}} \\ 
    DistMult \cite{ICLR_2015_Yang_DistMult}      & .241 & 254 & .155 & .263 & .419 & & .430 & 5110 & .390 & .440 & .490 \\
    ComplEx  \cite{ICML_2016_Trouillon_ComplEx} & .247 & 339 & .158 & .275 & .428 & & .440 & 5261 & .410 & .460 & .510 \\
    TuckER \cite{EMNLP_2019_Balazevic_TuckER} & .358 & -   & .266 & .394 & .544 & & .470 & -    & .443 & .482 & .526 \\
    ConvE  \cite{AAAI_2018_Dettmers_ConvE_WN18RR} & .325 & 244 & .237 & .356 & .501 & & .430 & 4187 & .400 & .440 & .520\\
    InteractE \cite{AAAI_2020_Vashishth_InteractE} & .354 & 172 & .263 & - & .535 & & .463 & 5202 & .430 & - & .528\\ 
    \textsc{ProcrustEs} \cite{NAACL_2021_Peng_ProcrustEs} & .345 & - & .249 & .379 & .541 & & .474 & - & .421 & \textbf{.502} & .569\\ 
    \hline
    \multicolumn{12}{l}{\textbf{GNN-based}} \\ 
    R-GCN \cite{ESWC_2018_Schlichtkrull_R-GCN}      & .248 & -   & .151 & - & .417 & & -  & -  & -  & - & -    \\
    KBGAT \cite{ACL_2019_Nathani_KBGAT}*     & .157 & 270 & - & - & .331 & & .412 & \textbf{1921} & - & - & .554 \\
    SACN \cite{AAAI_2019_Shang_SACN}         & .350 & -   & .260 & .390 & .540 & & .470 & -    & .430 & .480 & .540\\
    A2N \cite{ACL_2019_Bansal_A2N}         & .317 & -   & .232 & .348 & .486 &  & .450 & -    & .420 & .460 & .510\\
    CompGCN \cite{ICLR_2020_Vashishth_CompGCN}       & .355 & 197 & .264 & .390 & .535 & & .479 & 3533 & .443 & .494 & .546 \\
    \hline
      \textbf{CP-GNN} (ours)        & \textbf{.365} & 178 & \textbf{.276} & \textbf{.397} & \textbf{.554} & & \textbf{.482} & 3214 & \textbf{.447} & .492 & \textbf{.571}\\
    \bottomrule
  \end{tabular}
  \caption{
  Knowledge Graph Completion results on FB15k-237 and WN18RR dataset. H@1, H@3 and H@10 denote the metrics of Hits@1, Hits@3 and Hits@10 respectively. The best results are in \textbf{bold}. 
  CP-GNN achieves the SOTA performance in the overall consideration of five metrics on two datasets. 
  The results of TransE, RotatE, DistMult, ComplEx and ConvE are from \cite{ICLR_2019_Sun_RotatE}. 
  * means that the results of KBGAT are from \cite{ACL_2020_Sun_re-eval} because original results suffer from same score evaluation problem, which is discussed in section \ref{sec: exp setup}. 
  Other results are from the published paper.
  }
  \label{tab: kgc result}
\end{table*}

\subsection{Experimental Results of KGC Task}
\label{sec: kgc result}
Our baselines are selected from three categories which are \textbf{Translational Distance Models}: TransE \cite{NeurIPS_2013_Bordes_TransE}, RotatE \cite{ICLR_2019_Sun_RotatE}, PaiRE \cite{ACL_2021_Chao_PaiRE}; \textbf{Semantic Matching Models}: DistMult \cite{ICLR_2015_Yang_DistMult}, ComplEx \cite{ICML_2016_Trouillon_ComplEx}, TuckER \cite{EMNLP_2019_Balazevic_TuckER}, ConvE \cite{AAAI_2018_Dettmers_ConvE_WN18RR}, InteractE \cite{AAAI_2020_Vashishth_InteractE}, \textsc{ProcrustEs} \cite{NAACL_2021_Peng_ProcrustEs}; \textbf{GNN-based Models}: R-GCN \cite{ESWC_2018_Schlichtkrull_R-GCN}, KBGAT \cite{ACL_2019_Nathani_KBGAT}, SACN \cite{AAAI_2019_Shang_SACN}, A2N \cite{ACL_2019_Bansal_A2N}, CompGCN \cite{ICLR_2020_Vashishth_CompGCN}. 

The experimental results are demonstrated in Table \ref{tab: kgc result}, where we can see that proposed CP-GNN model obtains state-of-the-art results compared to current methods. On FB15k-237 dataset, CP-GNN obtains best results using four of five metrics, and on WN18RR dataset CP-GNN also achieves best with MRR, H@1 and H@10 metrics. For the MRR metric, which is an important indicator to describe the general ranking performance, CP-GNN attains best results on both two datasets, showing the simultaneous modeling of two semantic patterns encodes a more comprehensive knowledge representation for downstream task. For H@k metrics, CP-GNN achieves best performance on five terms across two datasets. The only exception is H@3 on WN18RR, while the result is also competitive. This shows that CP-GNN maintains a high prediction accuracy for top ranking entities. For the MR metric, CP-GNN also gives a competitive performance. Note that for the models PaiRE \cite{ACL_2021_Chao_PaiRE} and KBGAT \cite{ACL_2019_Nathani_KBGAT} with the best MR reporting, their other metric outcomes are not so as outstanding, and CP-GNN still achieves the better overall performance.

In addition, we observe that CP-GNN shows a comparatively better performance on FB15k-237 dataset than WN18RR. We consider such performance differences are caused by the query complexity characteristics of two datasets. In FB15k-237, there are high rate of queries with multiple answers (also known as 1-N relations from triple view), which demands higher modeling capacity to capture latent relevancy among answers, i.e. proximity pattern. While in WN18RR most queries only satisfy one answer (also known as 1-1 relations), implying WN18RR is an easier dataset that traditional relation pattern is sufficient to some extent. This can also be proven from the consistently better performance on WN18RR relative to FB15k-237 across most models. We think that proximity pattern plays a more important role in complex query scenarios like in FB15k-237, which will be further discussed in section \ref{subsec: N-type ablation}.

\section{Effectiveness Evaluation of Proximity Pattern}
In this section we attempt to answer following questions: \\ 
\textbf{Q1}. Does the employment of proximity pattern do improve the model performance for Knowledge Graph Completion task? (section \ref{subsec: pp ablation}) \\ 
\textbf{Q2}. How does the proximity pattern perform for data with different modeling complexity? (section \ref{subsec: N-type ablation})

\subsection{Ablation Study of Model Performance}
\label{subsec: pp ablation}
To evaluate the effect of proximity pattern for model performance on KGC task, we do the ablation study of removing the proximity pattern modeling module $\mathrm{G}_p$ in CP-GNN, and denote the remained architecture that only retains knowledge graph module $\mathrm{G}_r$ as CP-GNN (KG). Corresponding to the figure \ref{fig: model}, after relation pattern modeling the obtained embedding $\mathrm{E}_r$ will be directly input into the decoder as $\mathrm{E}_{enc}$. The relation embedding $R$ and its transformation will remain unchanged. The comparison results on FB15k-237 test set are summarized in Table \ref{tab: kgc ablation study}. We can observe the performance degeneration across all five metrics in CP-GNN (KG), which shows the limited capacity of only modeling relation pattern and the effectiveness of fusing into proximity pattern.

\begin{table}[t]
  \centering
  \begin{tabular}{lccccc}
    \toprule
    \multirow{2}{*}{\textbf{Models}} & \multicolumn{5}{c}{\textbf{FB15k-237}} \\
    \cline{2-6}
    & MRR & MR & H@1 & H@3 & H@10 \\
    \hline \hline
    \textbf{CP-GNN}     & \textbf{0.365} & \textbf{178} & \textbf{0.276} & \textbf{0.397} & \textbf{0.554}\\
    CP-GNN (KG) & 0.357 & 193 & 0.258 & 0.380 & 0.531 \\
    \bottomrule
  \end{tabular}
  \caption{Ablation study of proximity pattern for Knowledge Graph Completion task on FB15k-237 dataset.}
  \label{tab: kgc ablation study}
\end{table}

\subsection{Ablation Study on Data with Different Modeling Complexity}
\label{subsec: N-type ablation}
In this section, we will further probe into how does CP-GNN and CP-GNN (KG) perform in the subdivided data complexity scenarios. For each query-answer data, we decide its category based on its answer set size, which serves as an implication of modeling complexity. Intuitively, more answers need to simultaneously satisfy for one query, higher probability the conflicts may happen, which puts forward higher requirements for the model to capture the latent closeness relevancy among answer entities. In practice, for every $(h, r, t)$ we will count the satisfied entities in \textbf{train} set of query $(h, r, ?)$ and $(?, r, t)$ as its two categories. We denote the category as \textbf{N-type}.

We divide N-type into six ranges: N$=$0, N$=$1, 1$<$N$<=$10, 10$<$N$<=$100, 100$<$N$<=$500, N$>$500, representing the different complexity level. The statistics of N-type data in FB15k-237 and WN18RR is summarized in table \ref{tab: N-type statistics}. We can see that in FB15k-237 the proportion of complex data is obviously larger, where 10$<$N$<=$100 is 0.27 and 100$<$n$<=$500 is 0.13, and the counterparts in WN18RR are 0.09 and 0.04 respectively. This corresponds to our analysis in section \ref{sec: kgc result}, that FB15k-237 is a harder dataset so most models reveal a worse performance compared to WN18RR dataset. 

\begin{figure}[t]
  \centering
  \includegraphics[width=\columnwidth]{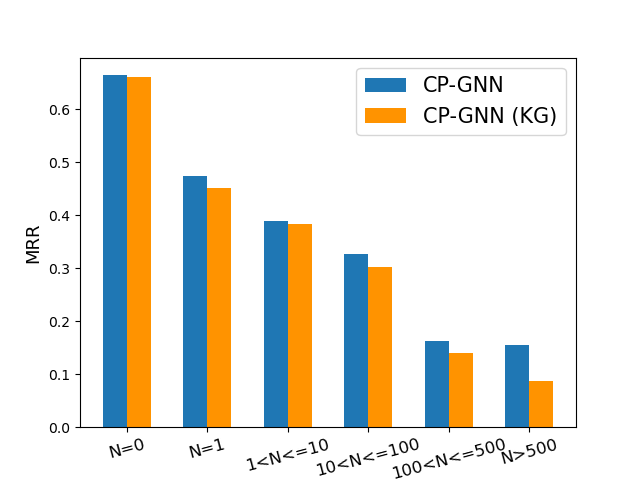}
  \caption{Ablation study of proximity pattern with regard to data with different modeling complexity in FB15k-237 dataset. The horizontal axis is the N-type range and vertical axis is MRR metric.}
  \label{fig: N-type ablation}
\end{figure}

We demonstrate the MRR result of CP-GNN and CP-GNN (KG) on different N-type ranges in figure \ref{fig: N-type ablation}. We can observe that CP-GNN outperforms or competes CP-GNN (KG) in all scenarios, and when N$>$10 the improvements are especially evident. This shows that the proximity pattern can effectively boost the modeling capacity for complex data. We consider this is because through proximity pattern, the potential answer entities will be modeled more close to each other, and the model can easily utilize observed facts to infer out unknown answers, like what illustrated in figure \ref{fig: proximity_inference}. 

\begin{table}[t]
  \centering
  \begin{tabular}{ccccc}
    \toprule
    \multirow{2}{*}{\textbf{N-type Range}} & \multicolumn{2}{c}{\textbf{FB15k-237}} & \multicolumn{2}{c}{\textbf{WN18RR}}\\
    \cline{2-5}
    & Num. & {Rate} & Num. & {Rate} \\
    \hline\hline
    N$=$0          & 6,881    & 0.17       & 2827    & 0.45  \\
    N$=$1       & 2,929    & 0.07       & 970    & 0.15   \\
    1$<$N$<=$10       & 11,368   & 0.28     & 1657    & 0.26 \\
    10$<$N$<=$100    & 10,965    & 0.27      & 581    & 0.09 \\
    100$<$N$<=$500    & 5,359    & 0.13      & 233    & 0.04  \\
    N$>$500        & 3,430    & 0.08       & 0    & 0.0  \\
    Total        & 40,932    & 1.0       & 6268    & 1.0\\
    \bottomrule
  \end{tabular}
  \caption{N-type statistics of FB15k-237 and WN18RR test set. N=0 means that there is no satisfied answer entities in train set. Because of each triple contributes $(h, r, ?)$ and $(?, r, t)$ two cases, the total number is double to the triple number.}
  \label{tab: N-type statistics}
\end{table}

\section{Conclusions}
In this work we explore the proximity pattern, a new semantic assumption of knowledge graph, which is able to help obtain more comprehensive knowledge embedding. Moreover, we design a Chained couPle-GNN (CP-GNN) architecture to fuse proximity pattern and original relation pattern in knowledge graph globally and deeply. Extensive experiments demonstrate the validity of proposed proximity pattern and the effectiveness of fused knowledge representations, especially in complex data scenarios. 

\bibliography{refs}
\clearpage
\appendix

\section{Weight Implementation Details}
\label{app: attention}
For the neighborhood aggregating mechanism of knowledge graph (equation \ref{eq: rp_agg}), $\alpha_{ij}^{(l)}$ describes the weight from source node $e_j$ to destination node $e_i$ for each layer, and the choices include:
\begin{enumerate}
  \item Prior Weight: The straight application of Prior Weight is an economical choice. For example, one can use $\alpha_{ij}^{(l)} = \frac{1}{d_i}$, the reciprocal of the outdegree of source nodes, implying that source message should be evenly spread to each destination node.
  \item GCN Weight: GCN Weight is derived from the Graph Convolutional Network \cite{ICLR_2017_Kipf_GCN}, which forms as $\alpha_{ij}=\frac{1}{\sqrt{d_i} \sqrt{d_j}}$, where $d_i$ is the degree of the node $i$. 
  \item Attention Weight: The Attention Weight is dynamically calculated in each layer, which can capture the different relative importance of each neighbor. Here we calculate the weight of each relational neighbor $(e_j, r_j)$ to node $e_i$ as:
  $$\alpha_{ij}^{(l)} = \mathrm{Softmax} \left( {(\mathbf{e}_i^{(l)})}^T \, \varphi(\mathbf{e}_j, \mathbf{r}_j) \right)$$
\end{enumerate}
We recommend the usage of attention weight to dynamically aggregate the neighbor information based on the importance, which is also shown performance improvement in our experiment (but in the cost of extra time and memory occupation).

\end{document}